\documentclass[sigconf]{acmart}

\usepackage{epsfig}
\usepackage{graphicx}
\usepackage{float}
\usepackage{amsmath}

\usepackage{amssymb}
\usepackage{booktabs}
\usepackage{array}
\usepackage{multirow}
\usepackage{url}

\AtBeginDocument{
  \providecommand\BibTeX{{
    \normalfont B\kern-0.5em{\scshape i\kern-0.25em b}\kern-0.8em\TeX}}}

\copyrightyear{2021}
\acmYear{2021}
\setcopyright{acmcopyright}
\acmConference[MM '21]{Proceedings of the 29th ACM International Conference on Multimedia}{October 20--24, 2021}{Virtual Event, China}
\acmBooktitle{Proceedings of the 29th ACM International Conference on Multimedia (MM '21), October 20--24, 2021, Virtual Event, China}
\acmPrice{15.00}
\acmDOI{10.1145/3474085.3475445}
\acmISBN{978-1-4503-8651-7/21/10}

\begin{document}
\fancyhead{}


\title{I2V-GAN: Unpaired Infrared-to-Visible Video Translation}


\author{Shuang Li}
\affiliation{%
  \institution{Beijing Institute of Technology}
  \state{Beijing}
  \country{China}
}
\email{shuangli@bit.edu.cn}

\author{Bingfeng Han}
\affiliation{%
  \institution{Beijing Institute of Technology}
  \state{Beijing}
  \country{China}
}
\email{bfhan@bit.edu.cn}

\author{Zhenjie Yu}
\affiliation{%
  \institution{Beijing Institute of Technology}
  \state{Beijing}
  \country{China}
}
\email{zjyu@bit.edu.cn}

\author{Chi Harold Liu}
\affiliation{%
  \institution{Beijing Institute of Technology}
  \state{Beijing}
  \country{China}
}
\email{liuchi02@gmail.com}
\authornote{Corresponding author.}

\author{Kai Chen}
\affiliation{%
  \institution{Yantai IRay Technologies Lt. Co.}
  \state{Shandong}
  \country{China}
}
\email{kai.chen@iraytek.com}

\author{Shuigen Wang}
\affiliation{%
  \institution{Yantai IRay Technologies Lt. Co.}
  \state{Shandong}
  \country{China}
}
\email{shuigen.wang@iraytek.com}


\begin{abstract}

Human vision is often adversely affected by complex environmental factors, especially in night vision scenarios. Thus, infrared cameras are often leveraged to help enhance the visual effects via detecting infrared radiation in the surrounding environment, but the infrared videos are undesirable due to the lack of detailed semantic information. In such a case, an effective video-to-video translation method from the infrared domain to the visible light counterpart is strongly needed by overcoming the intrinsic huge gap between infrared and visible fields. To address this challenging problem, we propose an infrared-to-visible (I2V) video translation method I2V-GAN to generate fine-grained and spatial-temporal consistent visible light videos by given unpaired infrared videos. Technically, our model capitalizes on three types of constraints: 1) adversarial constraint to generate synthetic frames that are similar to the real ones, 2) cyclic consistency with the introduced perceptual loss for effective content conversion as well as style preservation, and 3) similarity constraints across and within domains to enhance the content and motion consistency in both spatial and temporal spaces at a fine-grained level. Furthermore, the current public available infrared and visible light datasets are mainly used for object detection or tracking, and some are composed of discontinuous images which are not suitable for video tasks. Thus, we provide a new dataset for infrared-to-visible video translation, which is named IRVI. Specifically, it has 12 consecutive video clips of vehicle and monitoring scenes, and both infrared and visible light videos could be apart into 24352 frames. Comprehensive experiments on IRVI validate that I2V-GAN is superior to the compared state-of-the-art methods in the translation of infrared-to-visible videos with higher fluency and finer semantic details. Moreover, additional experimental results on the flower-to-flower dataset indicate I2V-GAN is also applicable to other video translation tasks. The code and IRVI dataset are available at \url{https://github.com/BIT-DA/I2V-GAN}.

\begin{figure}[!ht]
    \begin{tabular}{l}
    \includegraphics[height=1.8cm]{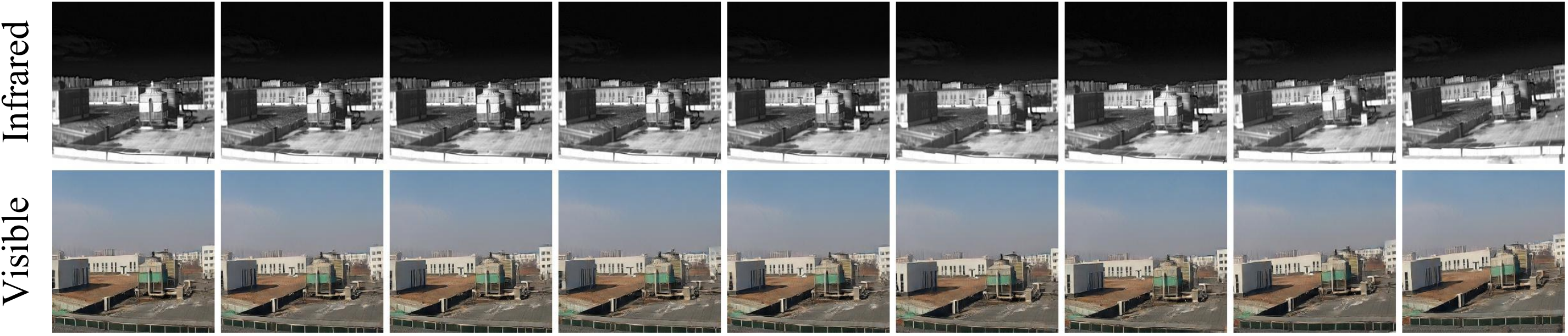} \\
    \includegraphics[height=1.8cm]{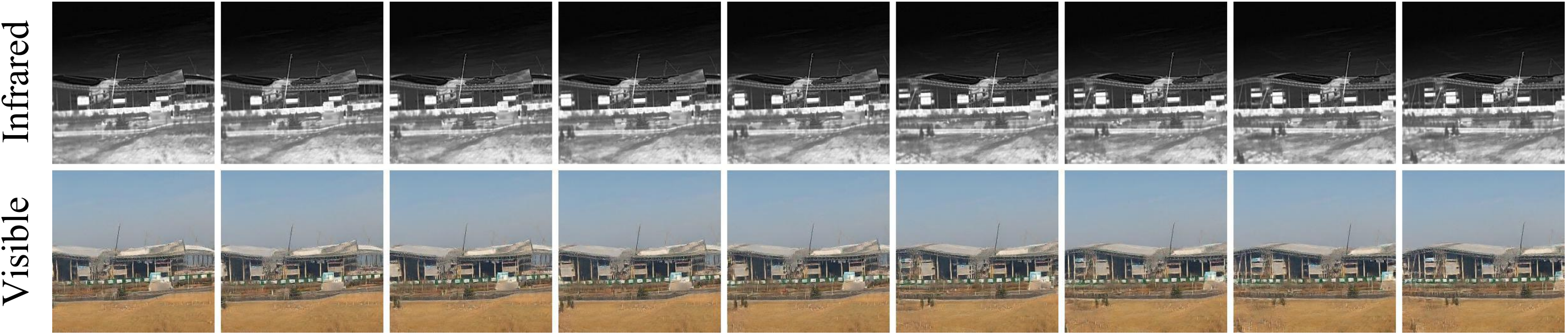} \\
    \includegraphics[height=1.8cm]{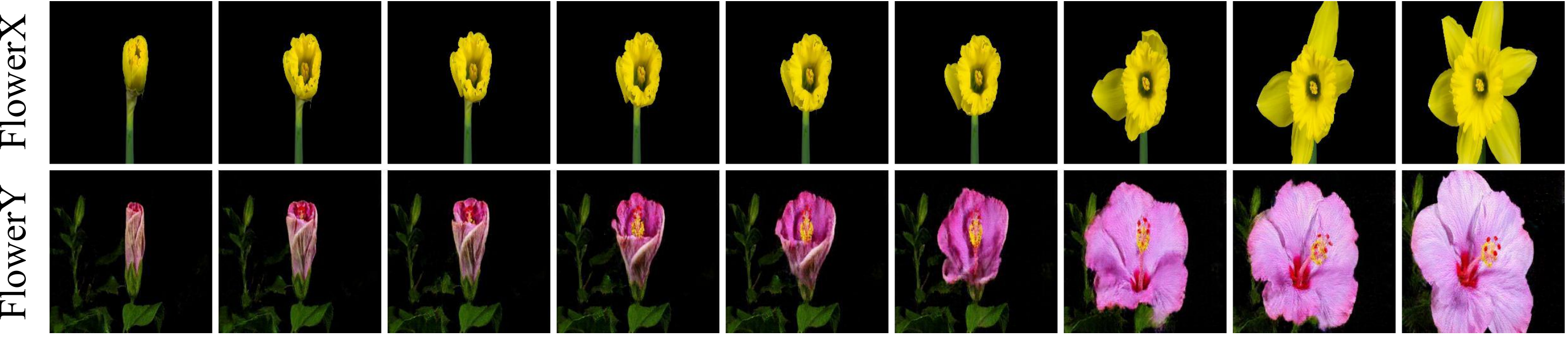}
    \end{tabular}
    \vspace{-1em}
    \caption{Given any two unpaired video collections $X$ and $Y$, our I2V-GAN learns to translate videos from the source domain $X$ to the target domain $Y$. The top four rows are infrared-to-visible results, and the other two rows are flower-to-flower results.}
    \label{fig:video translation examples}
    \vspace{-1em}
\end{figure}

\end{abstract}





\maketitle


\section{Introduction}

In real-world applications~\cite{intro_1, SSAN}, human vision is limited in nighttime scenarios and adverse weather conditions. Some vehicle navigation and monitoring systems using visible (VI) light cameras to enhance visual effects still obtain undesirable responses due to natural obstacles like different light conditions~\cite{intro_2}. Instead, infrared (IR) sensors gain advantages in capturing visual signals related to heat when visible-light cameras do not work well. However, compared with VI images, IR images have low color contrast and representation quality, which is difficult for people to recognize objects. In other words, VI images are easy to be recognized and contain more fine-grained semantic information but worse target contrast under bad luminance conditions, while IR images have better hot contrast but fewer environmental semantic details. Therefore, it is of great importance to generate VI color videos according to the corresponding collected IR visual signals. And the infrared-to-visible (I2V) video translation will have a broad application value in the multimedia and computer vision community, such as automatic driving and security fields. Unfortunately, the power of I2V video translation has yet to be fully unleashed in the previous works.

Classic I2V translation techniques are mainly based on image colorization methods, which can be divided into two main categories. The first type is the color morphing model via learning color mapping functions corresponding to the reference colors ~\cite{2002Color, 2006-manga, 2012Image, TCSVT, 2015A, 2016Fusion, 2016FACE}. But these methods are often time-consuming due to intensive manual interventions. For instance, ~\cite{2006-manga, TCSVT, 2012Image, 2016FACE} require pre-set reference color to paint on the gray-scale images or depend heavily on the manually selected reference color image. The converted images variate heavily according to the references. Another limitation is the confused orientation for VI images. ~\cite{2002gray, 2002Color, 2003color, 2007color} try to morph colors by finding filters or mapping functions from different color representation spaces. However, IR images are produced by heat, straight color morphing strategies are not effective to be applied. With the rapid development of deep learning methods ~\cite{li2020maximum,li2021faster,horsecolor, JADA} and transfer learning methods ~\cite{DICD, DRCNPDA} recently, image colorization methods delve into leveraging Generative Adversarial Networks (GANs) ~\cite{GAN} to colorize images automatically ~\cite{IR-DCGAN,2018colorize, 2018IR2VI, PCSGAN} by playing a min-max game. Although their translated images look similar to the visible light ones, they can not directly apply to video translation tasks owing to the temporal coherence among the adjacent frames.

As image-to-image translation methods advance rapidly, video-to-video translation goes one step further and grabs much attention ~\cite{RecycleGAN,MocycleGAN}. Specifically, Bansal et al. propose a data-driven video retargeting approach Recycle-GAN ~\cite{RecycleGAN} to consider both spatial and temporal constraints jointly based on the famous Cycle-GAN ~\cite{cycleGAN}. However, the motion changes between consecutive frames have not been fully exploited in Recycle-GAN. ~\cite{ReCoNet,2020Optical,MocycleGAN} utilize optical flow to maintain temporal consistency by considering the motion knowledge to alleviate the flickering artifacts effectively. Nevertheless, these methods rely heavily on optical flow extraction methods ~\cite{flownet,flownet2} and image warping algorithms with occlusion masks. Defects in each step could cause artifacts and blurry. The low efficiency of optical flow extraction for IR frames makes it ineffective for I2V video translation. As a consequence, the detailed semantic information in the converted VI videos are often unsatisfactory. Another solution to ensure temporal coherence for video translation is capitalizing on Recurrent Neural Networks (RNNs) ~\cite{MoCoGAN, UVIT}. To gather temporal and spatial information for the current frame, ~\cite{UVIT} applies forward and backward RNN units. Since the translation of each frame is related to all frames before and after it, these methods are time-consuming and hard to be applied for the real-time scenario.

To address the aforementioned problems, we propose an unpaired I2V video translation method I2V-GAN, which can generate spatial-temporal consistent videos with more elaborated semantic details and reach real-time performance at the inference phase inspired by Recycle-GAN ~\cite{RecycleGAN}. The main idea of Recycle-GAN is to simultaneously train a generator and a predictor to learn spatial and temporal consistency. The generator and predictor synthesize frames in the target domain according to the input source frames and the past synthesized frames. By applying cyclic losses similar to \cite{cycleGAN}, Recycle-GAN improves translation performance for videos compared to images. However, as analysis in ~\cite{RecycleGAN, MocycleGAN} and practical results of some I2V tasks, the cyclic constraints are not sufficient to guarantee the synthesized frames strongly continuous and usually suffer from severe flickering artifacts. Therefore, to conduct detailed pixel-to-pixel alignment, we introduce perceptual cyclic losses and two similarity losses across and within the domains to improve the translation performance. 

\begin{figure}[t]
    \centering
    \includegraphics[height=4cm]{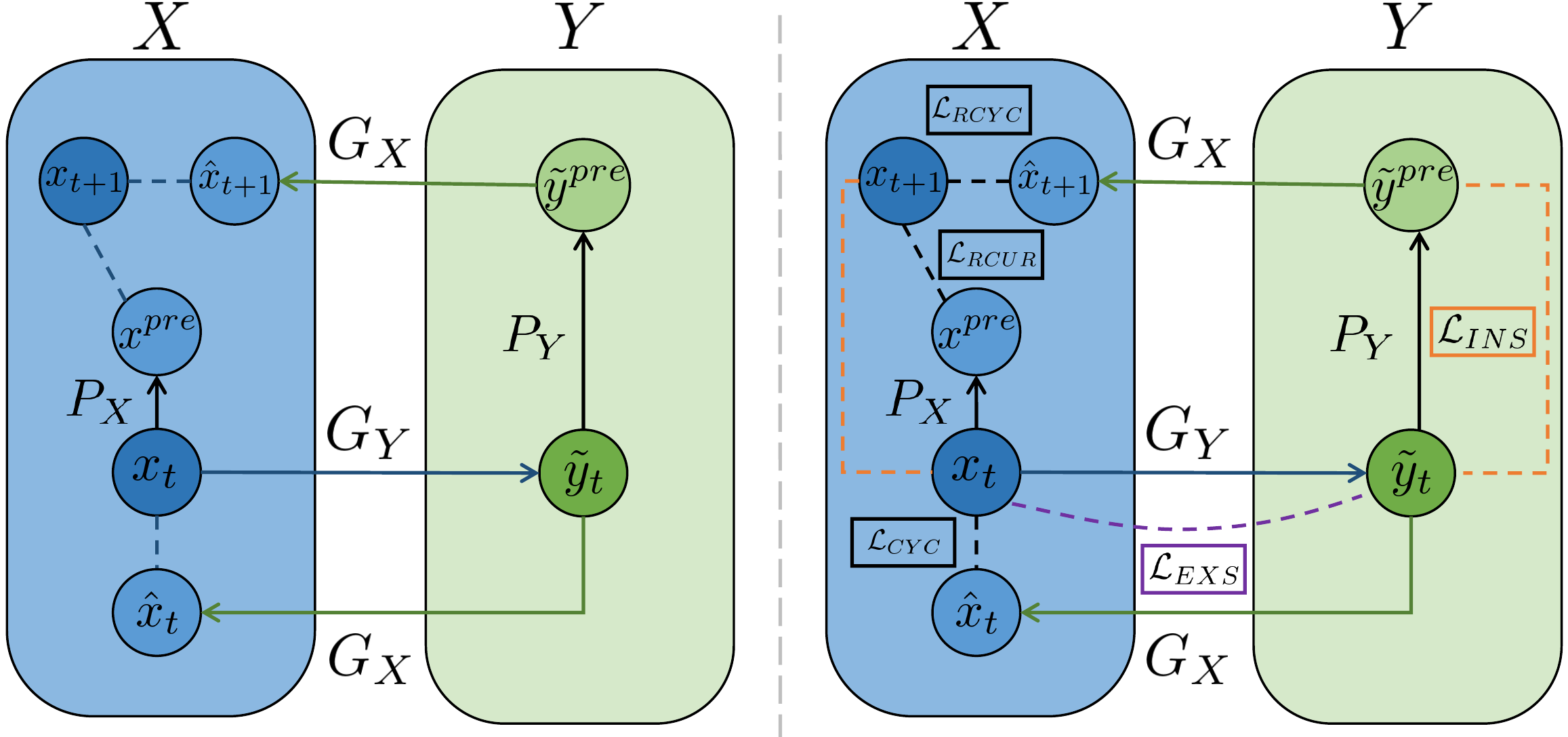}
    \vspace{-2em}
    \caption{Flow graph for Recycle-GAN (left) and I2V-GAN (right). The graph shows the translation flow of $X \rightarrow Y \rightarrow X$. The black dash lines in I2V-GAN are different perceptual cyclic constraints corresponding to Recycle-GAN. The orange and purple dash lines are internal and external similarity losses. More details are illustrated in section ~\ref{Section3}.}
    \label{fig:network_recycle}
    \vspace{-1em}
\end{figure}

\begin{figure*}[ht]
\centering
\includegraphics[height=7cm]{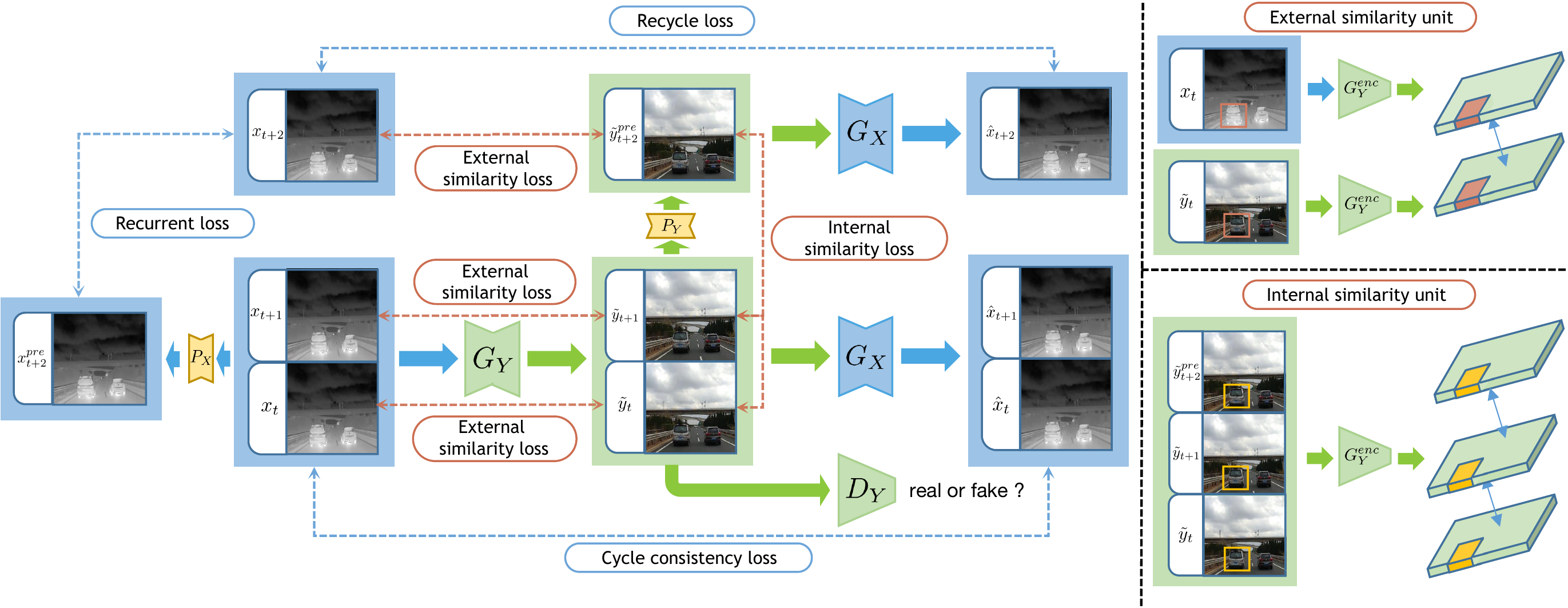}
\vspace{-2em}
\caption{The I2V-GAN network architecture with the flow of $X \rightarrow Y$. The opposite direction $Y \rightarrow X$ is similar. The frames on the \textcolor[RGB]{0,176,240}{blue} background is from the source domain, and the other frames on the \textcolor[RGB]{0,176,81}{green} background are expected to lay in the target domain. The network contains two generators ($G_{X}$ and $G_{Y}$) and two predictors ($P_{X}$ and $P_{Y}$) to synthesize frames across and within domain, respectively. Meanwhile, two discriminators ($D_{X}$ and $D_{Y}$) are to distinguish real frames from synthesized ones. Given three consecutive frames $x_{t}$, $x_{t+1}$, $x_{t+2}$ from domain $X$, we first synthesize frames $\tilde{y}_{t}, \tilde{y}_{t+1}$ by $G_{Y}$, which are further reconstructed to $\hat{x}_{t}, \hat{x}_{t+1}$ by $G_{X}$. Meanwhile, the synthesized frame $\hat{y}^{pre}_{t+2}$ by $P_{Y}$ with the former synthesized frames $\tilde{y}_{t}, \tilde{y}_{t+1}$ is also mapped to domain $X$ as $\hat{x}_{t+2}$ by $G_{X}$. As above, we optimize: 1) \emph{cycle consistency loss} and 2) \emph{recycle loss}. Moreover, in this direction, we simultaneously optimize $P_{X}$ according to $x_{t}$ and $x_{t+1}$, which named as 3) \emph{recurrent loss}. To further enhance the performance of video translation at image level and video level, we introduce 4) \emph{external similarity loss} and 5) \emph{internal similarity loss}.}
\vspace{-1em}
\label{fig:network achitecture}
\end{figure*}

To be specific, the perceptual loss is an additional constraint for each cyclic loss shown in Figure ~\ref{fig:network_recycle}. We apply a VGG loss network to optimize the feature extraction for the content and style. The optimization of style is combined with the Gram matrix of the representation, and content is related to the features extracted from certain layers of VGG. As such, the original constraints are equipped with fine-grained perceptual information from a pixel-wise perspective. Moreover, aiming to capture more semantic details in both domains, we propose an external similarity loss across domains to maximize the mutual information between the patches at the same location of the input frames and the synthesized frames. Besides, an internal similarity loss within the domain is introduced to keep the same motion variation degree of synthesized consecutive frames as the corresponding input frames. In this sense, the visual content in each frame can be transferred in a fine-grained way, and the spatial-temporal coherence is ensured to be realistic and consolidated simultaneously. The network flow of I2V-GAN is shown in Figure ~\ref{fig:network achitecture}. Our main purpose is to generate fine-grained transferred videos with high fluency and continuity.

Additionally and the same important as above, the current most-used public IR and VI datasets are unsuited for I2V video translation tasks. For example, ~\cite{OTCBVS} offers monitoring frames of still scenes, which have no motion variation over half of the time and lack diversity. ~\cite{FLIR} focuses on object detection and the image pairs are not strictly consecutive. ~\cite{VOT2019} is organized for object tracing and the length of each video clip is limited, which is not tailored to video translation evaluation. In this situation, we provide a new dataset named IRVI for IR and VI video translation. Our dataset contains 12 video clips for vehicle and monitoring scenarios. Both infrared and visible light videos could be apart into 24352 frames. More detailed descriptions and comparison will be illustrated in Section ~\ref{Section4}.

In summary, the contributions of our work are highlighted as below:

(1) To our best knowledge, our proposed I2V-GAN is the first end-to-end unpaired infrared-to-visible video translation network.

(2) To generate visible videos with higher fluency and finer semantic details, the improved cyclic constraints with content and style perceptual losses as well as external and internal similarity losses across and within domains are proposed.

(3) We provide a new dataset for infrared-to-visible video translation task named IRVI. Moreover, we present the benchmark of the IRVI for some state-of-the-arts and open-source methods.

\section{Related Work}

\textbf{Infrared-to-Visible Translation.}
Many vehicles navigation and monitoring systems apply infrared sensors to help enhance visual signals. However, the synthesis of IR images is related to heat, it is difficult for people to capture information. This motivates us to colorize IR images to the visible domain. Some basic transfer methods for I2V image translation, e.g., \cite{2015A, 2016Fusion, 2018IR2VI}, are proposed to translate IR images into gray-scale, not color ones. On the other hand, current I2V image translation works are mostly proposed for human face tasks ~\cite{zhang2018synthesis, damer2019cascaded}. Although ~\cite{IR-DCGAN, nyberg2019unpaired} transfer IR images to VI images, the color and quality are limited. Moreover, these methods cannot be directly applied to the I2V video translation task. Since it not only requires each video frame to look realistic but also should produce temporally coherent frames.

\textbf{Video-to-Video Translation.} 
The development of video-to-video translation is stimulated by the advance of image-to-image translation, which aims to learn a mapping to translate images from the source domain to the target domain. ~\cite{pix2pix, cycleGAN, starGAN, cartoonGAN} are proposed to solve different graphics tasks in image-to-image translation, e.g., semantic labels to photo, edges to photo, and photo to animation. In particular, Cycle-GAN ~\cite{cycleGAN} utilizes a cycle consistency loss to constrain output in the target domain under unpaired conditions. Soon after, ~\cite{pix2pixHD} and ~\cite{starGAN-v2} are proposed to improve the translation details. However, directly applying the image translation techniques to videos, the generated video will inevitably be affected by severe flickering artifacts. Besides, image translation only guarantees spatial consistency and does not involve temporal continuity. 

In order to effectively eliminate these issues, ~\cite{ReCoNet, vid2vid, vid2vid2, 2020Optical, MocycleGAN} utilize optical flow and image warping algorithms to maintain temporal consistency. For instance, Mocycle-GAN ~\cite{MocycleGAN} proposes the motion translation and consistency strategies based on Cycle-GAN to preserve the spatial and temporal consistency. ~\cite{vid2vid} relies heavily on labeled data and works in a supervised way, which is not suitable for I2V task. ~\cite{vid2vid2} is improved on the basis of \cite{vid2vid} and adapts for the few-shot manner. However, these methods are ineffective for I2V video translation due to the low efficiency of optical flow extraction under low contrast between consecutive frames. Inspired by Cycle-GAN, Recycle-GAN ~\cite{RecycleGAN} proposes to translate video via additional recurrent loss and recycle loss, enabling a spatial-temporal consistency. While as analysis in ~\cite{RecycleGAN, MocycleGAN} and results in some I2V video translation tasks, these constraints are not sufficient to make the synthesized frames strongly continuous as realistic visual videos. Moreover, these methods are all not tailored to I2V video translation, since how to guarantee the semantic details preservation is also crucial when given low contrastive infrared videos. 

In this paper, we take a step further by introducing perceptual cyclic losses and similarity losses for detailed I2V video translation, which can achieve higher fluency and finer semantic information. The perceptual cyclic losses aim to improve translation performance for both content and style, and the similarity losses could improve video consistency in both spatial and temporal spaces.

\section{The Proposed Method}\label{Section3}
\subsection{Notation and Problem Setting}

For two video collections $X=\{$$\textbf{x}$$\}$ in source domain and $Y=\{$$\textbf{y}$$\}$ in target domain, we denote the video $x$ in $\{$\textbf{x}$\}$ as ordered frames $\left\{{x_0}, {x_1}, ... , {x_t}\right\}$ and the video $y$ in $\{$\textbf{y}$\}$ as ordered frames $\left\{ {y_0}, {y_1}, ... , {y_s}\right\}$, respectively. In particular, ${x_t}$ denotes the $t$-th frame in video $x$ at time $t$. The goal of video-to-video translation is to learn two mapping functions between the source domain $X$ and the target domain $Y$. In more specific, $G_{Y}: X \rightarrow Y$ is a generator to translate video frames $\{{x_{t}}\}^{t=T}_{t=0}$ from domain $X$ to domain $Y$ as $\{\tilde{y}_{t}\}^{t=T}_{t=0}$ and $G_{X}: Y \rightarrow X$ is a generator to translate video frames $\{{y}_{s}\}^{s=S}_{s=0}$ from domain $Y$ to domain $X$ as $\{\tilde{x}_{s}\}^{s=S}_{s=0}$. Our main purpose is to translate IR video to VI video with spatial-temporal consistency and make it in line with people's cognitive habits. Moreover, a discriminator $D_{Y}$ is applied to distinguish real frames $\{y_{s}\}^{s=S}_{s=0}$ from translated frames $\{\tilde{y}_{t}\}^{t=T}_{t=0}$. Similarly, another discriminator $D_{X}$ distinguishes real frames $\{{x_{t}}\}^{t=T}_{t=0}$ from $\{\tilde{x}_{s}\}^{s=S}_{s=0}$. In the rest of section ~\ref{Section3}, we mainly illustrate our method in the direction of $X \rightarrow Y$.

\subsection{Adversarial Constraint}
For GAN ~\cite{GAN} based methods ~\cite{cycleGAN, starGAN, RecycleGAN, MocycleGAN}, the generators and discriminators are adversarially trained to improve their mutual performance. Specifically, given synthesized frames $\{\tilde{y}_{t}\}^{t=T}_{t=0}$ generated by $G_{Y}$ and real frames $\{{y}_{s}\}^{s=S}_{s=0}$, the discriminator $D_{Y}$ is trained to correctly distinguish the fake synthesized frames. Meanwhile, the generator is trained to synthesis high-quality frames in order to fool the discriminator. Therefore, the adversarial network could generate frames similar to targets via the following loss:
\begin{equation}
\begin{aligned}
\mathcal{L}_{ADV} = \sum_{s}logD_{Y}(y_{s}) + \sum_{t}log(1-D_{Y}(G_{Y}(x_{t}))).
\end{aligned}
\label{eq1}
\end{equation}

\subsection{Recycle-GAN Revisit}
\textbf{Cycle Consistency Loss.}
Cycle-GAN ~\cite{cycleGAN} proposes cycle consistency loss for unpaired image-to-image translation. Specifically, the generator $G_Y$ synthesizes a fake frame $\tilde{y}_t$ corresponding to the input frame $x_t$, the other generator $G_X$ should reconstruct frame $x_t$ from $\tilde{y}_t$, which can denote as: $\hat{x}_t = G_{X}(G_{Y}(x_t))$, and the reconstruction loss is then formulated as:
\begin{equation}
\begin{aligned}
\mathcal{L}_{rec}(G_{X}, G_{Y}) = \Vert x_{t} - G_{X}(G_{Y}(x_{t})) \Vert_{1}.
\end{aligned}
\label{eq2}
\end{equation}

Recycle-GAN ~\cite{RecycleGAN} takes a step further based on Cycle-GAN. Besides cycle consistency loss in the spatial space, it introduces two additional losses for translation consistency in the temporal space.

\textbf{Recurrent Loss.}
Firstly, it applies a temporal predictor $P_{X}$ to predict future frame $x^{pre}_{t+1}$ based on an ordered frames sequence $\{x_0, x_1, ..., x_t\}$. The frame $x^{pre}_{t+1}$ should be the same as $x_{t+1}$:
\begin{equation}
\begin{aligned}
\mathcal{L}_{recurrent}(P_{X}) = \sum_{t}\Vert x_{t+1} - P_{X}(x_{0:t}) \Vert^{2},
\end{aligned}
\label{eq3}
\end{equation}
where we write ordered frames sequence $\{x_0, x_1, ..., x_t\}$ as $x_{0:t}$.

\textbf{Recycle Loss.}
Along with the above-mentioned predictor, a richer constraint across the domain and time is proposed:

\begin{equation}
\begin{aligned}
\mathcal{L}_{recycle}(G_{X}, G_{Y}, P_{Y}) = \sum_{t}\Vert x_{t+1} - G_{X}(P_{Y}(G_{Y}(x_{0:t}))) \Vert^{2},
\end{aligned}
\label{eq4}
\end{equation}
where $G_{Y}(x_{0:t}) = (G_{Y}(x_0), G_{Y}(x_1), ..., G_{Y}(x_t))$. This approach finds a way to constrain the generators $G_{X}$, $G_{Y}$ and the predictor $P_{Y}$ in both spatial and temporal spaces. As for implementation, the predicted future frame $x_{t}$ is related to the past frames $x_{t-2}$ and $x_{t-1}$.

\subsection{Perceptual Loss}
Actually, Cycle-GAN has achieved satisfying style translation performance, which only focuses on image-to-image translation. This paper concentrates on video-to-video translation, we propose an additional perceptual loss for fine-grained translation at the image level. Our perceptual loss works from two aspects of content and style. The content perceptual loss calculates the feature discrepancy between synthesized frames and real frames. The style perceptual loss computes the gram loss between synthesized frames and real frames at layer $l$ of the VGG network. Accordingly, our perceptual loss for ${X} \rightarrow {Y}$ direction is formulated as:

\vspace{-1em}
\begin{equation}
\mathcal{L}^{\phi_l}_{CON}(G_{X}, G_{Y}) = \frac{1}{C_{l}H_{l}W_{l}} \sum_t\Vert \phi_{l}(x_t) - \phi_{l}(G_{X}(G_{Y}(x_{t}))) \Vert^2,
\label{eq5}
\end{equation}

\vspace{-1em}
\begin{equation}
\begin{aligned}
\mathcal{L}^{\phi_l}_{STY}(G_{X}, G_{Y}) = \Vert Gram_{\phi_l}(x_t) - Gram_{\phi_l}(G_{X}(G_{Y}(x_{t}))) \Vert^{2}_F,
\end{aligned}
\label{eq6}
\end{equation}

\vspace{-1em}
\begin{equation}
\begin{aligned}
\mathcal{L}_{PCP}(G_{X}, G_{Y}) = \mathcal{L}_{CON}(G_{X}, G_{Y}) + \mathcal{L}_{STY}(G_{X}, G_{Y}),
\end{aligned}
\label{eq7}
\end{equation}
where $\phi_{l}$ represents $l$-th layer of the VGG network, $C_l$ represents the channels of the feature map and $H_l$, $W_l$ represent the size of the feature map, respectively at layer $l$. $Gram(\cdot)$ is calculation of Gram matrix. The losses for ${Y}\rightarrow{X}$ direction is similar.

We apply this perceptual loss $\mathcal{L}_{PCP}$ as additional constraint in $cycle$ $consistency$ $loss$, $recurrent$ $loss$ and $recycle$ $loss$. Each refined loss can be formulated as below: 

\begin{equation}
\begin{aligned}
\mathcal{L}_{CYC}(G_{X}, G_{Y}) = &\sum_t\Vert x_{t} - G_{X}(G_{Y}(x_{t})) \Vert_{1} \\
&+ \mathcal{L}_{PCP}(x_{t}, G_{X}(G_{Y}(x_{t}))),
\end{aligned}
\label{eq8}
\end{equation}

\begin{equation}
\begin{aligned}
\mathcal{L}_{RCUR}(P_{X}) = &\sum_t(\Vert x_{t+1} - P_{X}(x_{0:t}) \Vert_{1} \\
&+ \mathcal{L}_{PCP}(x_{t+1}, P_{X}(x_{0:t}))),
\end{aligned}
\label{eq9}
\end{equation}

\begin{equation}
\begin{aligned}
\mathcal{L}_{RCYC}(G_{X}, G_{Y}, P_{X}) = & \sum_t(\Vert x_{t+1} - G_{X}(P_{Y}(G_{Y}(x_{0:t}))) \Vert_{1} \\
&+ \mathcal{L}_{PCP}(x_{t+1}, G_{X}(P_{Y}(G_{Y}(x_{0:t}))))),
\end{aligned}
\label{eq10}
\end{equation}
here we alter $L2$ $loss$ to $L1$ $loss$ for more precise translation. As for implementation, we keep the same three-consecutive-frames training mechanism as Recycle-GAN. The network structure and training flow are presented in Figure ~\ref{fig:network achitecture}.

\subsection{Similarity Loss}
In practical scenarios, although the perceptual cyclic losses help the network learn video translation in a more detailed way, there still exist mismatches of color between consecutive frames, which makes it not good enough for real application. In this circumstance, we introduce the $InfoNCE$ $loss$ to our network, which helps detailed translating by maximizing mutual information. This noise contrastive estimation idea is to mining the relationship between two kinds of signal patches (normally notated as ``query'' and ``positive''). Meanwhile, contrast the $query$ to other patches (referred to as ``negatives''). We apply this similarity estimation across domains and within domain to maximize spatial-temporal consistency, which named as $external$ $similarity$ $loss$ and $internal$ $similarity$ $loss$, respectively.

\begin{figure}[t]
    \centering
    \includegraphics[height=4cm]{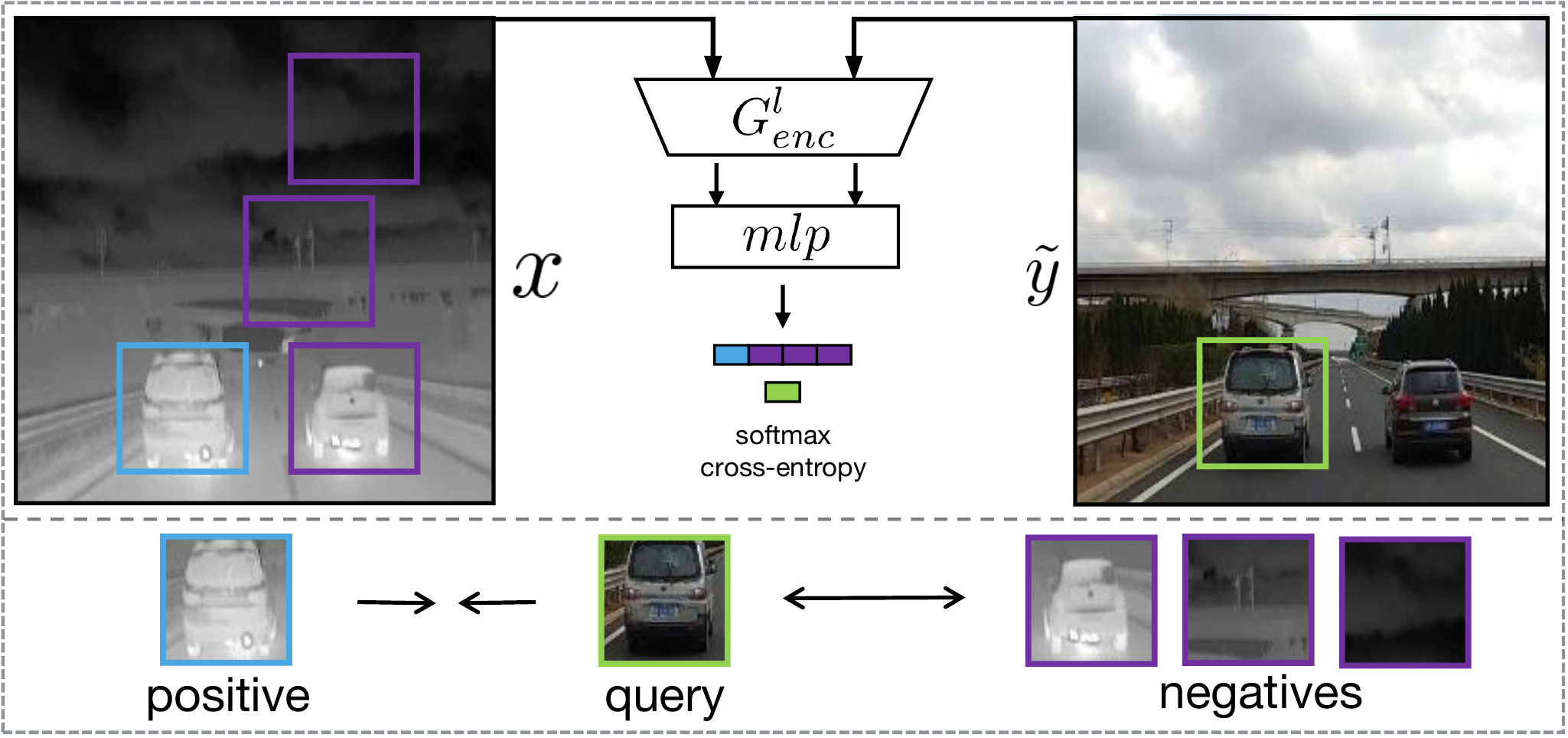}
    \vspace{-1em}
    \caption{External similarity loss. The frame $x$ is the input from domain $X$ and $\tilde{y}$ is the corresponding synthesized frame in domain $Y$. For each synthesized patch in $\tilde{y}$, it should correlate to the same spatial location in $x$ as much as possible, i.e., the synthesized van patch as ``query'' should be close to the ``positive'' and increase the distance from ``negatives''.}
    \vspace{-1em}
    \label{fig:ExS}
\end{figure}

\textbf{External Similarity Loss.}
The noise contrastive estimation ~\cite{Representation} can be regarded as an $(N+1)$-way classification problem. The input frames and synthesized frames go through the encoder at each layer $l$ of the generator, then a two-layer $MLP$ ~\cite{MLP} helps to project selected patches to a shared embedding space. The $query$, $positive$ and $N$ $negatives$ are mapped into $K$-dimensional vectors as $v$, $v^{+}$ and $N$ $v^{-}$, which can be formulated by cross-entropy as:
\begin{equation}
\begin{aligned}
\mathcal{L}_{NCE}(v, v^{+}, v^{-}) = -\log	\left[\frac{exp({v}\cdot{v^{+}}/\tau)}{exp({v}\cdot{v^{+}}/\tau) + \sum^{N}_{n}exp(v\cdot{v^{-}_{n}}/\tau)}\right],
\end{aligned}
\label{eq11}
\end{equation}
where $\tau$ is temperature scale parameter. Moreover, for each $query$ patch in synthesized frames, the corresponding $positive$ patch is the same location in the input frames and the other $negatives$ patches are random selected from inputs for $external$ $similarity$ $loss$, as shown in Figure~\ref{fig:ExS} and we formulate it as:

\begin{equation}
\begin{aligned}
\mathcal{L}_{EXS}(X) = \sum^{L}_{l}\sum^{S}_{s}\mathcal{L}_{NCE}(\tilde{z}^{s}_{l}, z^{s}_{l}, z^{S\backslash{s}}_{l}),
\end{aligned}
\label{eq12}
\end{equation}
where $\tilde{z}^{s}_{l}$ represents the output features at spatial location $s$ of $MLP$ at layer $l$ for synthesized frame. ${z}^{s}_{l}$ represents corresponding positive features and $z^{S\backslash{s}}_{l}$ represents the other negative features. The purpose is to restrict translation between corresponding patches, i.e. the white van should be more closely associated to the van patch, rather than the car and sky patches.

\begin{figure}
    \centering
    \includegraphics[height=4.8cm]{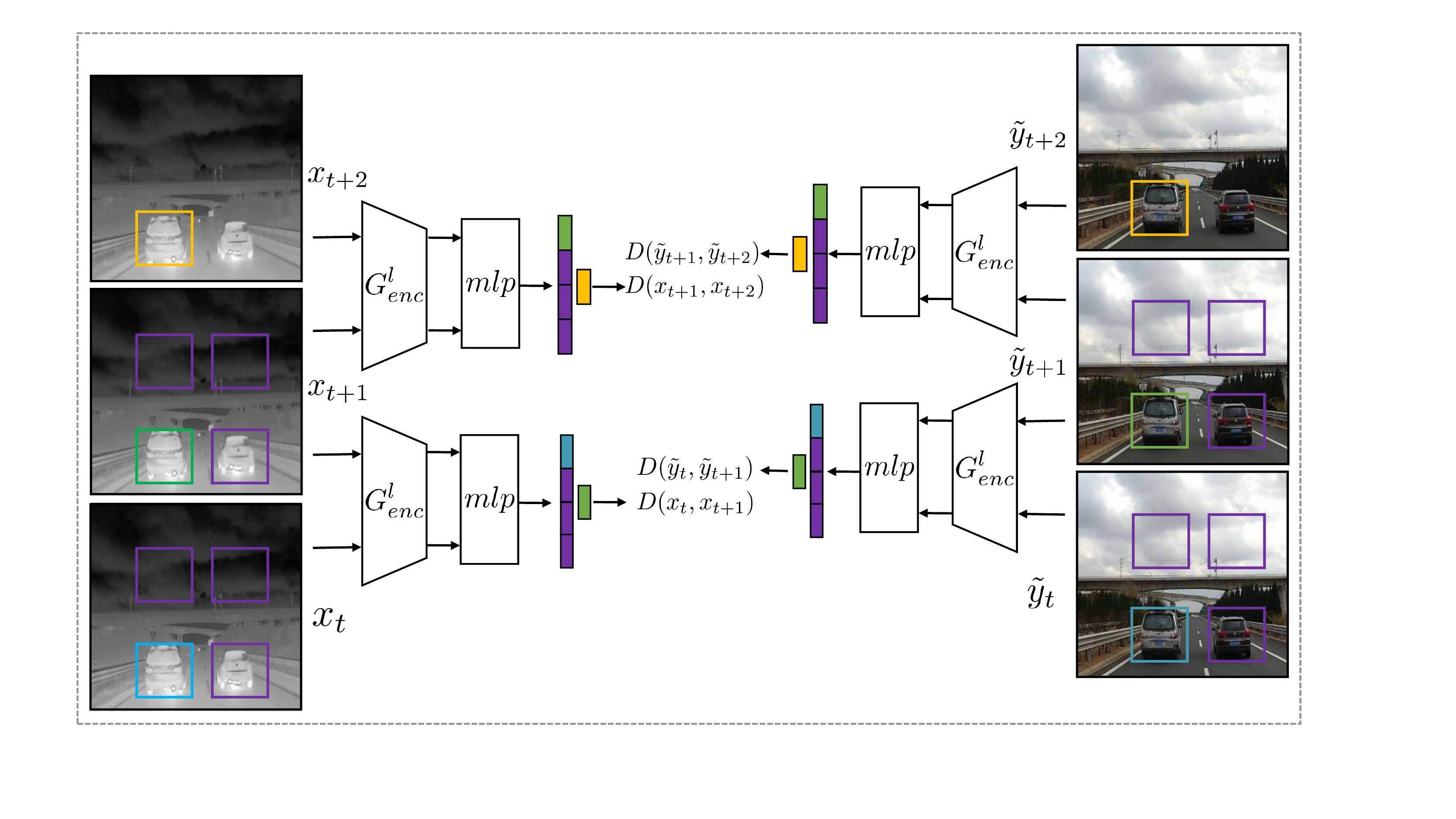}
    \caption{Internal similarity loss. The frames $x$, $x_{t+1}$, $x_{t+2}$ are inputs from the domain $X$. $\tilde{y}$, $\tilde{y}_{t+1}$ and $\tilde{y}_{t+2}$ are corresponding synthesized frames by I2V-GAN. We first compute the mutual information via \emph{NCE} between consecutive frames <$x$, $x_{t+1}$> and <$x_{t+1}$, $x_{t+2}$>, then treat the ratio of the two as the standard motion variation degree. After we got the standard ratio, we further compute the motion degree ratio of <$\tilde{y}$, $\tilde{y}_{t+1}$> and <$\tilde{y}_{t+1}$, $\tilde{y}_{t+2}$>. By restricting the two motion variation degrees we could improve spatial-temporal consistency for synthesized video.}
    \vspace{-1em}
    \label{fig:InS}
\end{figure}

\textbf{Internal Similarity Loss.}
Inventively, we utilize mutual information to represent the similarity between consecutive frames and propose $internal$ $similarity$ $loss$ within the domain.

Intuitively, classification confidence associate to the appearance of the object: the more appearance of the object, the higher degree of confidence. Combined with the conversion of noise contrastive estimation for mutual information, we can further regard the variation of $InfoNCE$ as a motion variation degree between consecutive frames. In order to measure this variation, we first compute the discrepancy between consecutive input frames and treat it as a criterion. In more specific, the mapped feature vectors at layer $l$ of $MLP$ represent current spatial-temporal information at level $l$. We contrast the feature vectors of two consecutive frames at the same layer by calculating their $InfoNCE$ $loss$. Since we selected $L$ layers of interests, we could obtain an $L$-dimensional vector $v_{\left \langle x_{t}, x_{t+1} \right \rangle}$ to present discrepancy of $L$ levels. Meanwhile, the corresponding synthesized frames could also obtain an $L$-dimensional vector $v_{\left \langle \tilde{y}_{t}, \tilde{y}_{t+1} \right \rangle}$. Our goal is to constrain the degree of motion between the generated consecutive frames to be consistent with the corresponding real frames. Thus, we simultaneously compute vector $v_{\left \langle x_{t+1}, x_{t+2} \right \rangle}$ and $v_{\left \langle \tilde{y}_{t+1}, \tilde{y}_{t+2} \right \rangle}$, as shown in Figure~\ref{fig:InS} and we formulate our $internal$ $similarity$ $loss$:

\begin{equation}
\begin{aligned}
\mathcal{D}_{SIM}(x_{t}, x_{t+1}, x_{t+2}) = \frac{v_{\left \langle {x}_{t+1}, {x}_{t+2} \right \rangle}}{v_{\left \langle {x}_{t}, {x}_{t+1} \right \rangle}},
\end{aligned}
\label{eq13}
\end{equation}

\begin{equation}
\begin{aligned}
&\mathcal{L}_{INS}(X)  = \\
& \sum_{t}(1 - \frac{\mathcal{D}_{SIM}(x_{t}, x_{t+1}, x_{t+2}) \cdot \mathcal{D}_{SIM}(\tilde{y}_{t}, \tilde{y}_{t+1}, \tilde{y}_{t+2})}{\Vert \mathcal{D}_{SIM}(x_{t}, x_{t+1}, x_{t+2}) \Vert \cdot \Vert \mathcal{D}_{SIM}(\tilde{y}_{t}, \tilde{y}_{t+1}, \tilde{y}_{t+2}) \Vert}),
\end{aligned}
\label{14}
\end{equation}
where $\mathcal{D}_{SIM}(\cdot,\cdot,\cdot)$ represents the variation degree of three consecutive frames. Moreover, since our goal is to constrain this variation degree, we need a rigorous standard from input frames and keep the corresponding relationship between relevant patches. The $negatives$ are selected in the same spatial order for each optimization step, rather than random selection as in $external$ $similarity$ $loss$.

\subsection{Overall Optimization}
In summary, our final objective function is organized as below:

\begin{equation}
\begin{aligned}
\mathcal{L} = & \mathcal{L}_{ADV} + \lambda_{1}\cdot\mathcal{L}_{CYC} + \lambda_{2}\cdot\mathcal{L}_{RCUR} + \lambda_{3}\cdot\mathcal{L}_{RCYC} \\
& + \lambda_{4}\cdot\mathcal{L}_{EXS} + \lambda_{5}\cdot\mathcal{L}_{INS},
\end{aligned}
\label{eq15}
\end{equation}
where we set perceptual cyclic losses tradeoff parameters $\lambda_{1}=\lambda_{2}=\lambda_{3}$. $\lambda_{4}$ and $\lambda_{5}$ are $external$ $similarity$ $loss$ and $internal$ $similarity$ $loss$ tradeoff parameters, respectively.


\section{IRVI Dataset}\label{Section4}

In general, infrared sensors include two types: near-infrared camera and long-wave infrared camera. The radiation band of the human body is within the range of the latter one. Therefore, the long-wave infrared camera is more in line with the requirements of recognition, security, and vehicle driving scenes. We use the long-wave type equipments to collect data.

\vspace{-1em}
\begin{table}[!ht]
    \centering
    \small
    \caption{The structure of IRVI}
    \vspace{-1em}
    \begin{tabular}{| m{13mm}<{\centering} | m{7mm}<{\centering} | m{12mm}<{\centering} | m{12mm}<{\centering} | m{10mm}<{\centering} | m{10mm}<{\centering} |}
        \hline
        \multicolumn{2}{|c|}{\textbf{SUBSET}} & \textbf{TRAIN} & \textbf{TEST} & \multicolumn{2}{c|}{\textbf{TOTAL FRAME}} \\
        \hline
        \multicolumn{2}{|c|}{Traffic} & 17000 & 1000 & \multicolumn{2}{c|}{18000}\\
        \hline
        \multirow{5}{*}{Monitoring}
        & sub-1 & 1384 & 347 & 1731 & \\
        & sub-2 & 1040 & 260 & 1300 & \\
        & sub-3 & 1232 & 308 & 1540 & 6352\\
        & sub-4 & 672  & 169 & 841 & \\
        & sub-5 & 752  & 188 & 940 & \\
        \hline
    \end{tabular}
    \vspace{-1em}
    \label{tab:Dataset subsets.}
\end{table}

\vspace{-1em}

\begin{table}[!ht]
    \centering
    \small
    \caption{Datasets Comparision Information}
    \vspace{-1em}
    \begin{tabular}{|m{30mm}<{\centering}|m{11mm}<{\centering}|m{10mm}<{\centering}|m{20mm}<{\centering}|}
        \hline
        \textbf{NAME} & \textbf{FRAME} & \textbf{CLIP} & \textbf{TASK} \\
        \hline
        IRVI & 24352 & 12 & video translation \\
        \hline
        VOT2019 (RGBTIR) & 20083 & 60 & object tracking \\
        \hline
        FLIR & 4224 & 1 & \multirow{2}{*}{object detection} \\
        \cline{1-3}
        KAIST (DAY ROAD) & 16176 & 9 & \\
        \hline
    \end{tabular}
    \vspace{-1em}
    \label{tab:Datasets comparision}
\end{table}

\begin{figure}
    \centering
    \begin{tabular}{cc}
      \includegraphics[scale=0.13]{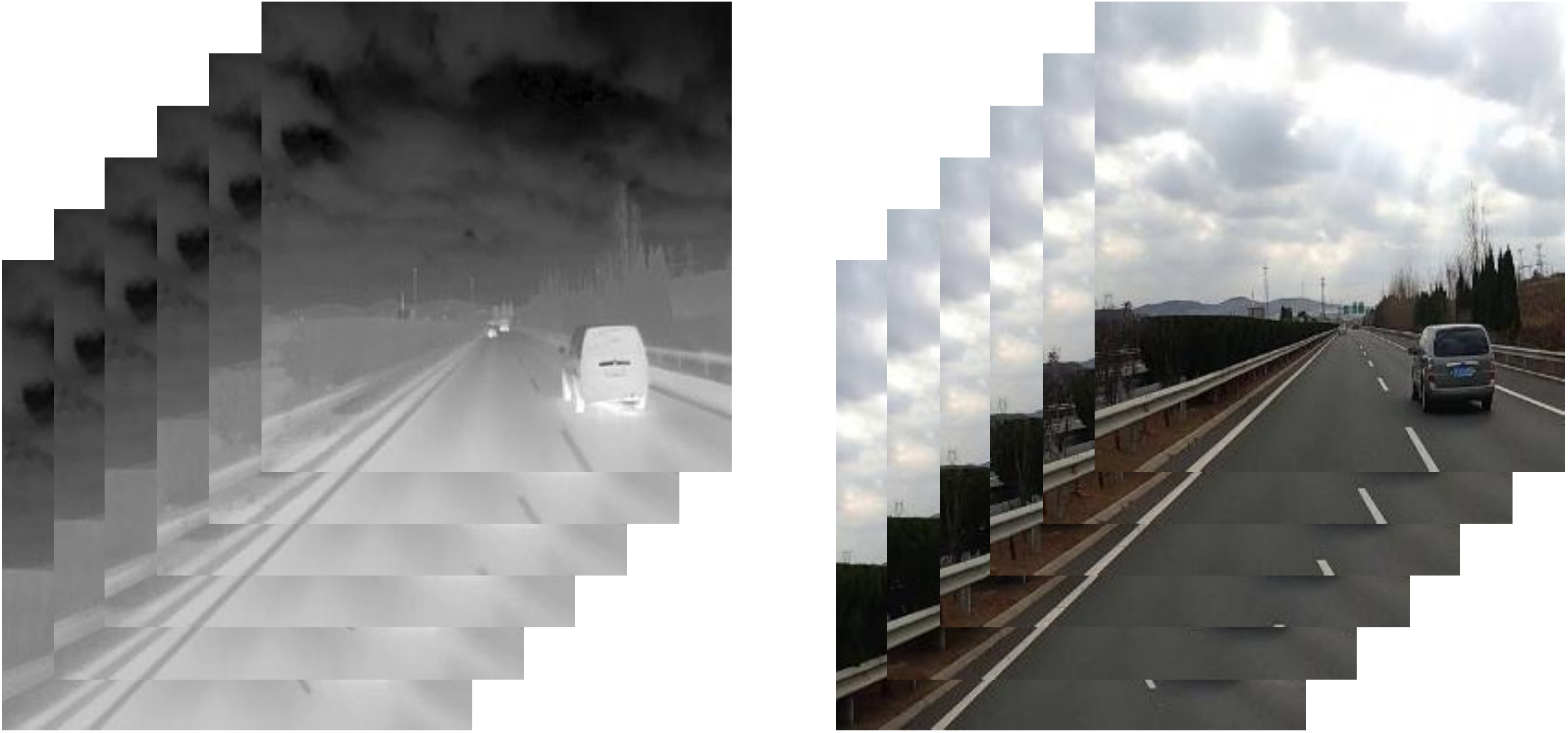} &
      \includegraphics[scale=0.13]{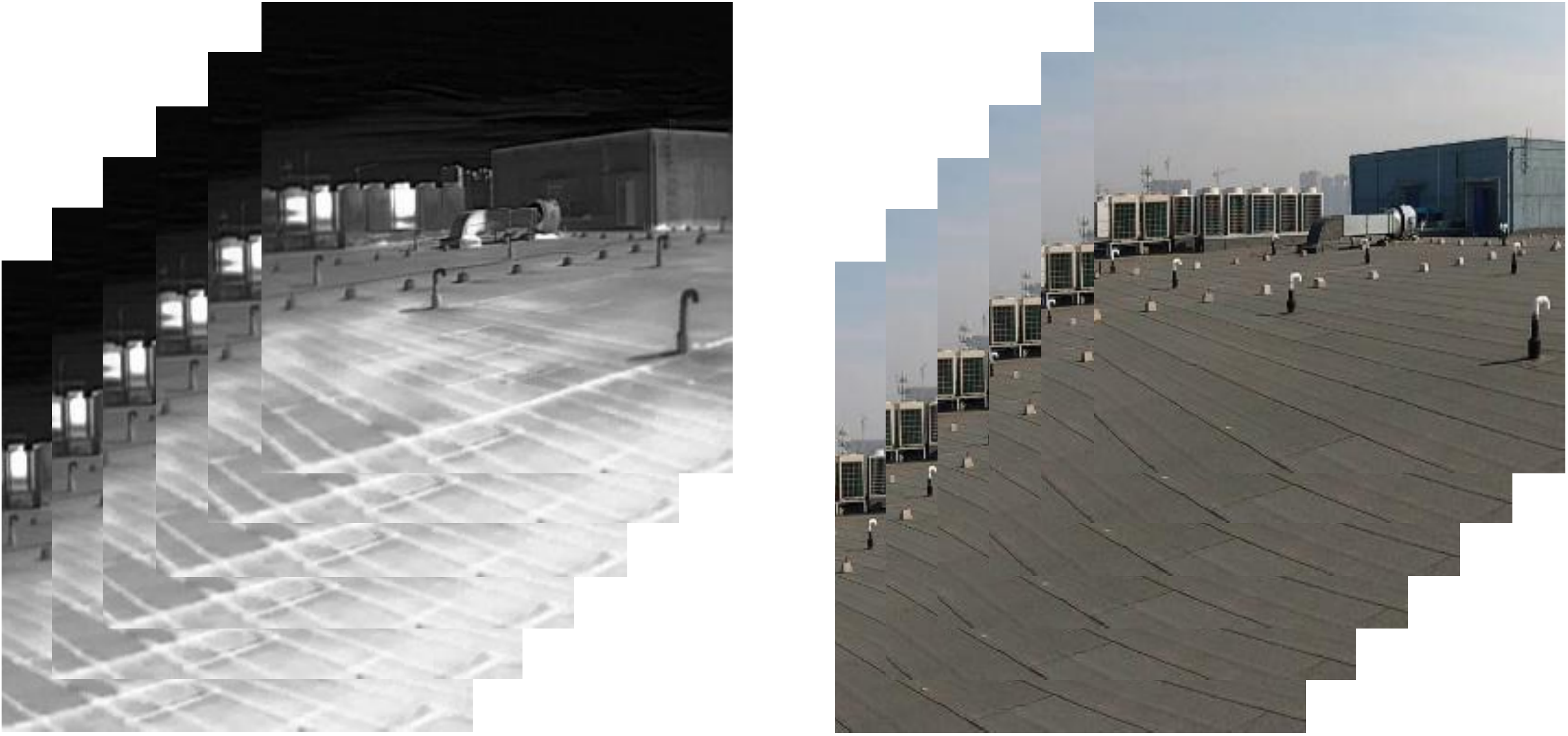} \\
      (a) & (b) \\
      \includegraphics[scale=0.13]{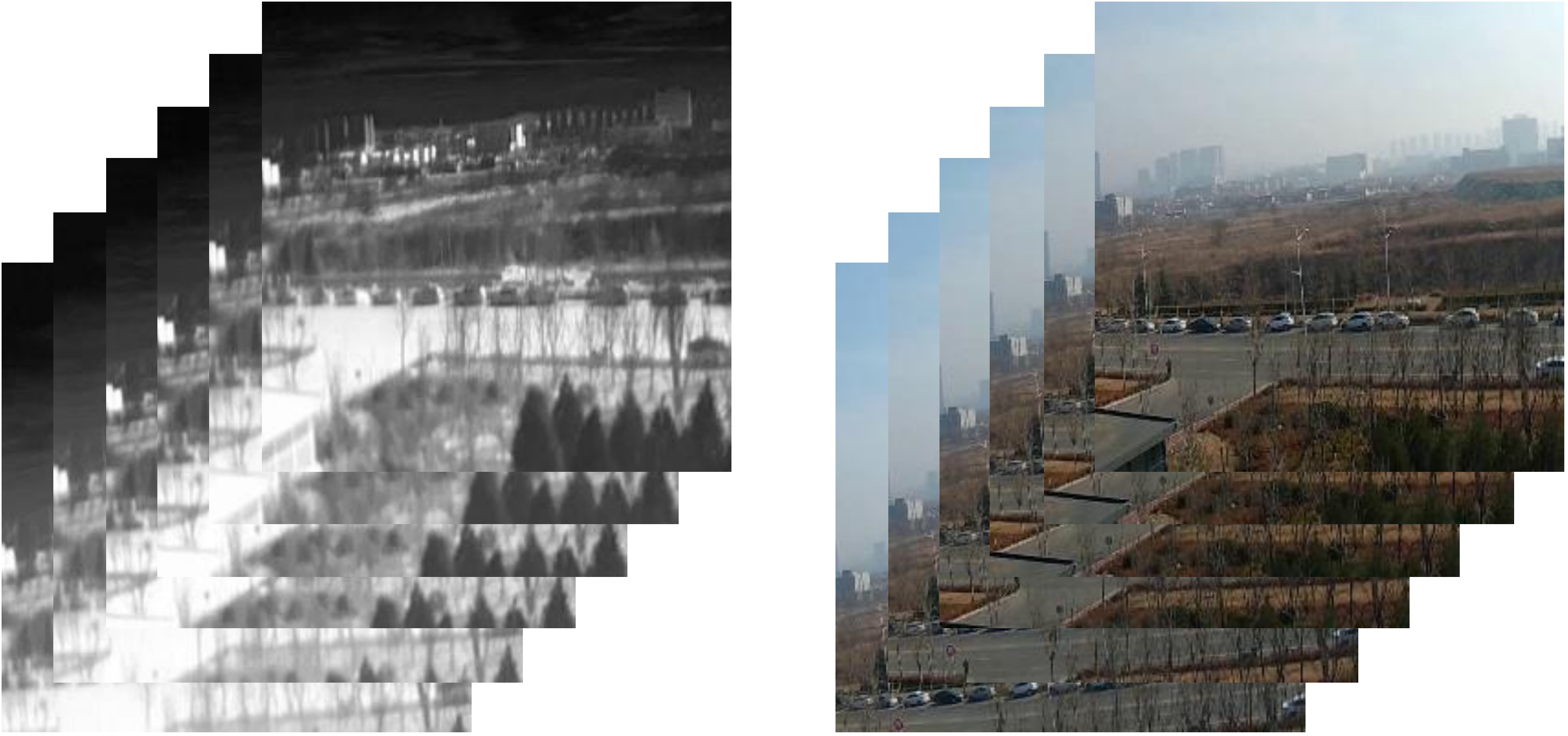} &
      \includegraphics[scale=0.13]{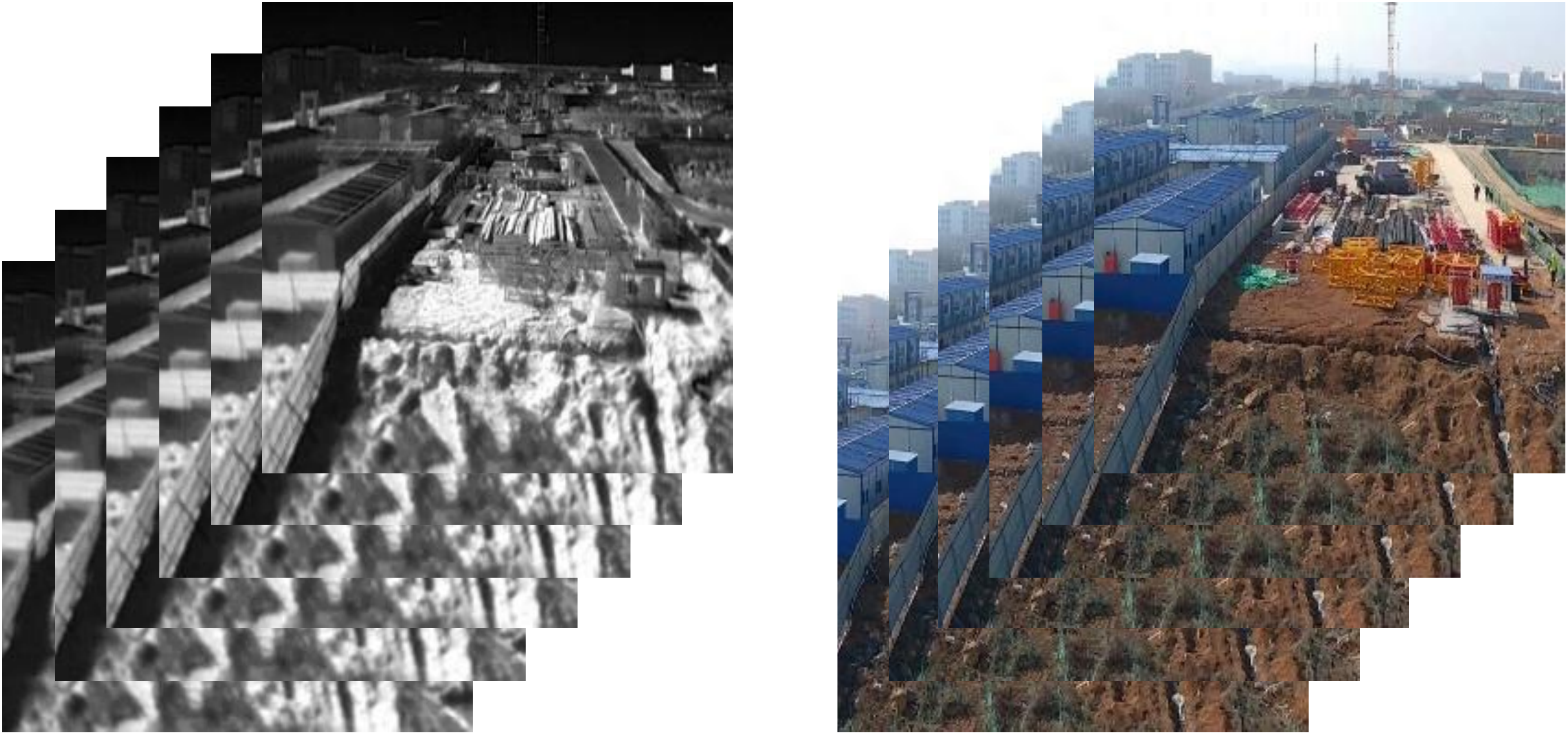} \\
      (c) & (d) \\
      \includegraphics[scale=0.13]{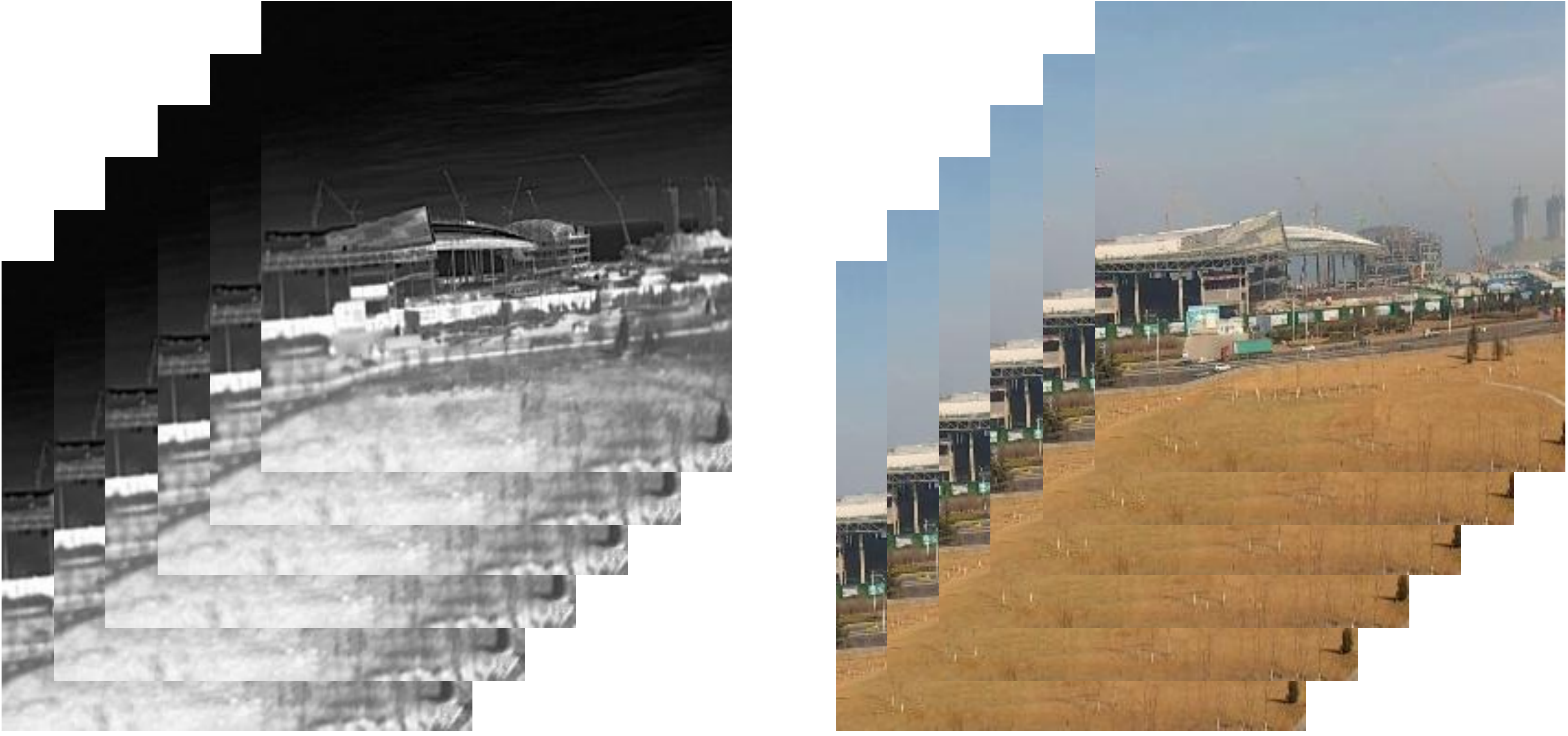} &
      \includegraphics[scale=0.13]{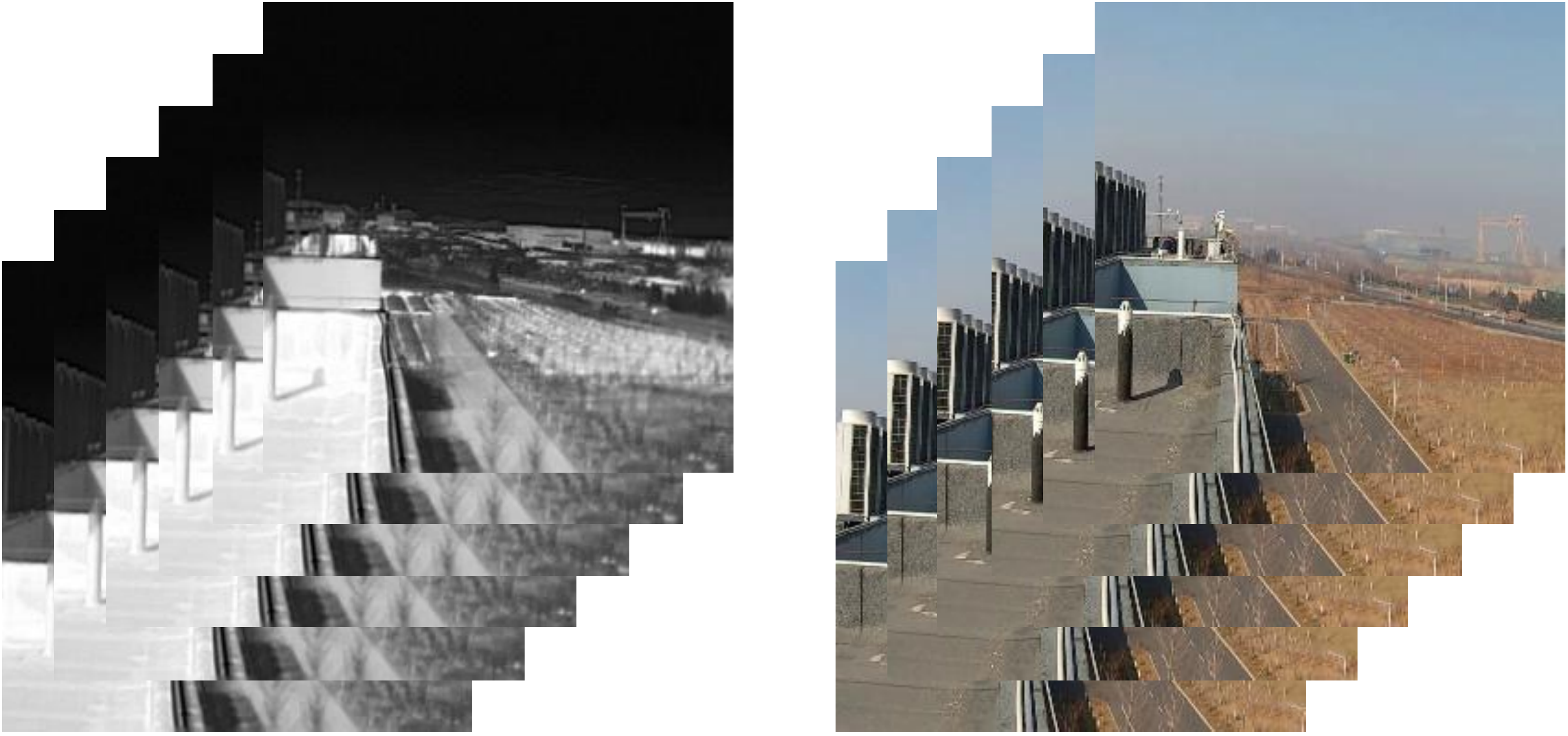} \\
      (e) & (f) \\
    \end{tabular}
    \vspace{-1em}
    \caption{Examples for IRVI. We present 6 consecutive frames from each subset of IRVI. In particular, (a) is from Traffic and (b)$\sim$(f) are from 5 subsets of Monitoring. Infrared on the left and corresponding visible light on the right.}
    \vspace{-1em}
    \label{fig:data example.}
\end{figure}

The dataset collects video streams through a binocular infrared color camera (DTC equipment) and performs scene alignment on the two domains at the hardware level. In different periods, the infrared and visible light data of traffic and monitoring scenes are obtained through vehicle-mounted and fixed-point brackets ways. Examples for each scene are shown in Figure \ref{fig:data example.}. Each example includes infrared frames and visible light frames. The composition and quantity of the dataset are detailed in the Table \ref{tab:Dataset subsets.}. 

We comprehensively select three public available infrared and visible light video datasets for comparison, which are VOT2019-RGBTIR ~\cite{VOT2019}, FLIR ~\cite{FLIR} and KAIST ~\cite{KAIST} in day road scene. The detailed comparison information is shown in Table ~\ref{tab:Datasets comparision}.

Specifically, VOT2019-RGBTIR provides 60 video clips, while the duration and quality are varied. Although FLIR offers other infrared and visible light image pairs besides the 4224 frames listed in Table ~\ref{tab:Datasets comparision}, they are inconsecutive and not counted in as videos. KAIST is the largest dataset among the three. It contains 6 subsets for training and another 6 subsets for testing. All the subsets are focus on traffic scene differs from place and time. Our IRVI dataset contains 6 clips for training and another 6 clips for testing. All our video clips are continuous as shown in Figure ~\ref{fig:data example.}. On the other hand, the other three datasets collect for object tracking and detection. Our dataset IRVI is collected for I2V video translation, which contains limited luminance scenes. Moreover, IRVI incorporates monitoring scenes besides traffic, which makes it practical for real-world applications.


\section{Experiments}
\subsection{Datasets and Setup}
\textbf{IRVI Dataset.} Our experiments are mainly based on IRVI dataset. This dataset consists of two parts: traffic and monitoring, as introduced in Section ~\ref{Section4} . The resolution of each video is scaled to 256 × 256. This translation task aims to colorize infrared video to visible light video as much similar as real color videos. Meanwhile, the translation results should eliminate the impact of adverse environmental factors, and within the scope of human cognition.

\textbf{Flower Video Dataset.} This dataset records the life cycle of different flowers through time-lapse photography, depicting the blooming or fading without any sync. The resolution of each video is 256 × 256. We evaluate the translation between different types of flowers as Recycle-GAN ~\cite{RecycleGAN} and Mocycle-GAN ~\cite{MocycleGAN}. This translation task aims to align two flowers simultaneously bloom or fade.

\textbf{Implementation Details.} We implement I2V-GAN network on Pytorch ~\cite{pytorch}. In detail, we use a Resnet-based generator ~\cite{residualblock}, which is a PatchGAN structure encoder-decoder. To embed the features extracted from each layer of the encoder, we apply a two-layer MLP with 256 units after encoding. In all experiments, we set the tradeoff parameters $ \lambda_{1}=\lambda_{2}=\lambda_{3}=10 $ for perceptual cyclic losses, and $\lambda_{4}=0.1$, $\lambda_{5}=40$ for $\mathcal{L}_{EXS}$ and $\mathcal{L}_{INS}$, respectively. During the training, the batch size is set as 1 to achieve better performance.

\begin{figure}[!t]
    \centering
    \includegraphics[height=8cm]{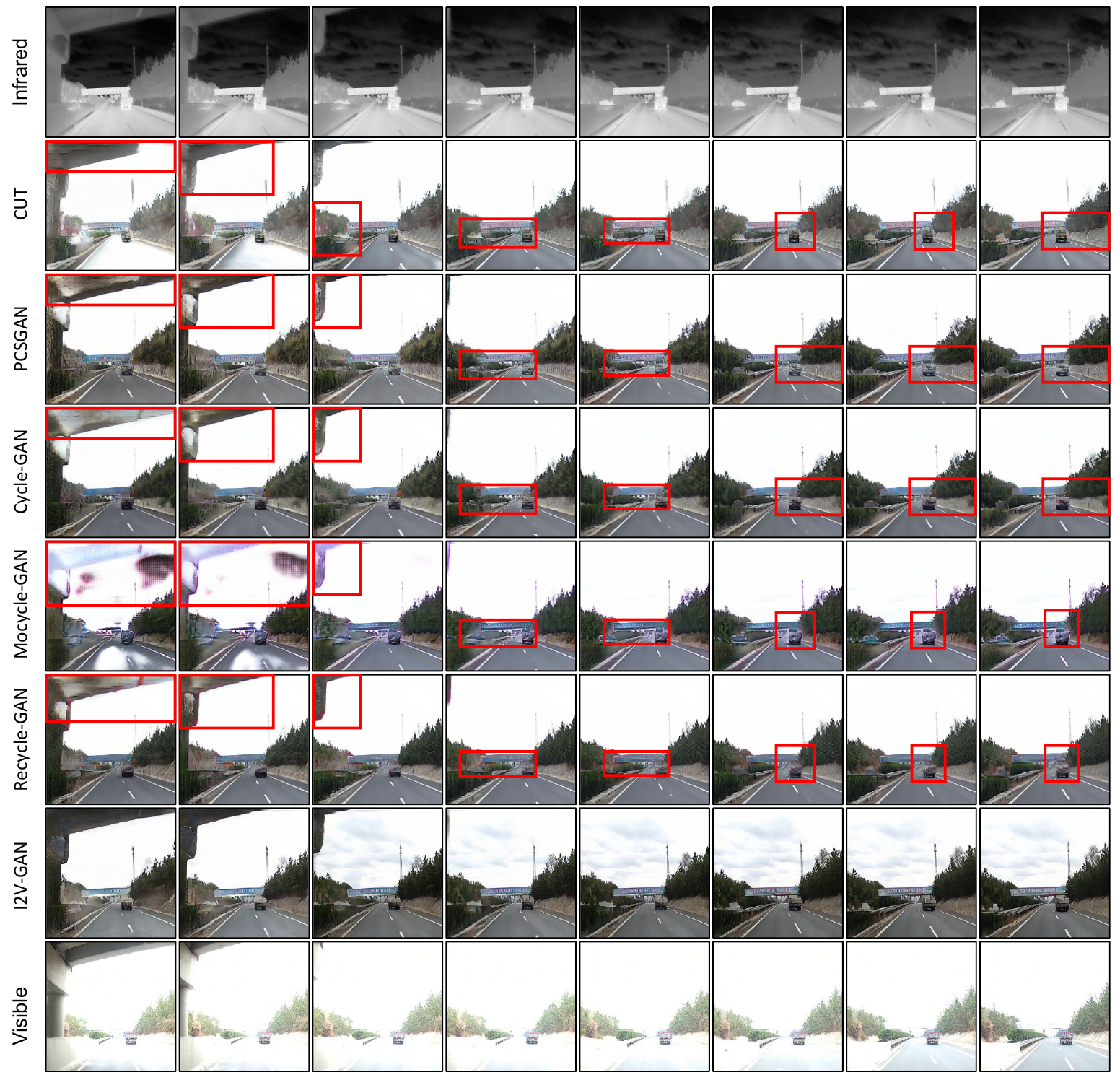}
    \vspace{-1em}
    \caption{I2V traffic samples. In this scene, the visible light camera is exposed when the vehicle passes under the bridge. Our goal is to eliminate this adverse via translating infrared frames and make them consistent with human cognition.}
    \vspace{-1em}
    \label{fig:expose}
\end{figure}

\subsection{Evaluation Metrics} For infrared-to-visible video translation and flower-to-flower translation tasks, we first evaluate Fréchet Inception Distance~(\textbf{FID})~\cite{FID} at image level. FID calculates the distance between the real frames and the synthesized frames in the feature space. And the feature representations are extracted from the inception network~\cite{FID_network}. The lower the value indicates that the distribution of the generated frames is closer to the real distribution. Then we evaluate Peak Signal-to-Noise Ratio ~(\textbf{PSNR}) and Structural Similarity ~(\textbf{SSIM}) from the pixel and structure levels, respectively. PSNR is generally regarded as an indicator of colorization methods to express translation quality. The higher value of PSNR indicates less distortion of the images. SSIM reveals the structure information of the objects carried by the interdependence of pixels. The higher value of SSIM means the more similar of the two comparison objects. Moreover, since FID, PSNR and SSIM could only represent performance to a certain extent and not fully suitable for video performance evaluation, we conduct an additional user study for application feedback as ~\cite{RecycleGAN, MocycleGAN, vid2vid, vid2vid2}.


\begin{table*}[!ht]
    \centering
    \renewcommand{\multirowsetup}{\centering}
    \caption{Fréchet Inception Distance for different translation methods. Lower is better.}
    \vspace{-1em}
    \label{tab:FID}
    \small
    \begin{tabular}{| m{23mm}<{\centering} | m{15mm}<{\centering} | m{15mm}<{\centering} | m{15mm}<{\centering} m{15mm}<{\centering} m{15mm}<{\centering} m{15mm}<{\centering} m{15mm}<{\centering} m{15mm}<{\centering}|}
    \hline
    \multirow{3}{*}{\textbf{Method}} & \multicolumn{8}{c|}{\textbf{FID ${\downarrow}$}} \\
    \cline{2-9}
     & \multirow{2}{*}{\textbf{Flower}} & \multirow{2}{*}{\textbf{Traffic}} & \multicolumn{6}{c|}{\textbf{Monitoring}} \\
    \cline{4-9}
     & & & sub-1 & sub-2 & sub-3 & sub-4 & sub-5 & all \\
    \hline
    CUT & 0.7861 & 0.5739 & 1.4348 & 0.6731  & 2.2294  & 1.0553 & 2.1973 & 1.0893 \\
    PCSGAN & 0.5680 & 0.6436 & \textbf{1.0963} & 0.7315 & 2.0843 & 1.1034 & 1.6808 & 0.8912 \\
    Cycle-GAN & 0.8164 & 0.6714 & 1.4027 & 0.8056 & 2.1497 & 1.0359 & 1.6266 & 0.8792 \\
    MoCycle-GAN & 0.7135 & 0.7911 & 1.5556 & 0.9847 & 2.5013 & 1.1040 & 2.1171 & 1.0515\\
    Recycle-GAN & 0.6306 & 0.5255 & 1.6680 & 0.7521 & 2.0387 & 1.2959 & 1.8518 & 1.0609 \\
    I2V-GAN & \textbf{0.4891} & \textbf{0.4425} & 1.4840 & \textbf{0.5905} & \textbf{1.7916} & \textbf{0.9189} & \textbf{ 1.6015} & \textbf{0.8715} \\
    \hline
    \end{tabular}
    \vspace{-1em}
\end{table*}

\begin{table*}[!ht]
    \centering
    \renewcommand{\multirowsetup}{\centering}
    \caption{Peak Signal-to-Noise Ratio and Structural Similarity for different translation methods. Larger is better.}
    \vspace{-1em}
    \label{tab:PSNRSSIM}
    \small
    \begin{tabular}{| m{23mm}<{\centering} |m{15mm}<{\centering} | m{15mm}<{\centering} | m{15mm}<{\centering} m{15mm}<{\centering} m{15mm}<{\centering} m{15mm}<{\centering} m{15mm}<{\centering} m{15mm}<{\centering}|}
    \hline
    \multirow{3}{*}{\textbf{Method}} & \multicolumn{8}{c|}{\textbf{PSNR/SSIM ${\uparrow}$}} \\
    \cline{2-9}
     & \multirow{2}{*}{\textbf{Flower}} & \multirow{2}{*}{\textbf{Traffic}} & \multicolumn{6}{c|}{\textbf{Monitoring}} \\
    \cline{4-9}
     & & & sub-1 & sub-2 & sub-3 & sub-4 & sub-5 & all \\
    \hline
    CUT & 9.27/0.33 & 16.86/0.56 & \textbf{15.87}/0.45 & 20.83/0.50 & 13.13/0.41 & 19.06/0.53 & 14.13/\textbf{0.15} & 17.00/0.43 \\
    PCSGAN & 9.43/0.34 & 15.32/0.55 & 15.18/0.48 & 20.49/0.49 & 13.37/0.40 & 18.52/0.52 & \textbf{14.52}/0.14 & 17.19/0.43 \\
    Cycle-GAN & 9.65/0.31 & 14.87/0.54 & 15.78/0.48 & 19.43/0.50 & 14.02/0.41 & 18.43/0.52 & 14.15/0.14 & 17.14/0.43 \\
    MoCycle-GAN & 9.39/0.37 & 15.60/0.56 & 14.99/0.43 & 19.39/0.50 & 11.51/0.40 & 18.84/0.54 & 13.83/\textbf{0.15} & 17.09/0.43 \\
    Recycle-GAN & 9.41/0.37 & 16.84/0.56 & 14.64/0.44 & 20.32/0.49 & 13.15/0.43 & 18.28/0.54 & 13.13/0.14 & 16.34/0.43 \\
    I2V-GAN & \textbf{10.58}/\textbf{0.40} & \textbf{17.02}/\textbf{0.60} & 14.81/\textbf{0.51} & \textbf{21.20}/\textbf{0.52} & \textbf{14.11}/\textbf{0.47} & \textbf{19.26}/\textbf{0.59} & 13.96/\textbf{0.15} & \textbf{17.30}/\textbf{0.46} \\
    \hline
    \end{tabular}
    \vspace{-1em}
\end{table*}

\begin{table*}[t]
    \caption{Ablation study and further investigation experimental results.}
    \vspace{-1em}
    \label{tab:ablation study}
    \small
    \begin{tabular}{|m{30mm}<{\centering} | m{15mm}<{\centering} | m{25mm}<{\centering} | m{15mm}<{\centering} | m{25mm}<{\centering} | m{15mm}<{\centering} | m{25mm}<{\centering}| }
        \hline
        \multirow{2}{*}{\textbf{Method}} & \multicolumn{2}{c|}{\textbf{Flower}} & \multicolumn{2}{c|}{\textbf{Traffic}} & \multicolumn{2}{c|}{\textbf{Monitoring} - all} \\
        \cline{2-7}
         & \textbf{FID} ${\downarrow}$ & \textbf{PSNR/SSIM} ${\uparrow}$ & \textbf{FID} ${\downarrow}$ & \textbf{PSNR/SSIM} ${\uparrow}$ & \textbf{FID} ${\downarrow}$ & \textbf{PSNR/SSIM} ${\uparrow}$ \\
        \hline
        Mocycle-GAN & 0.7135 & 9.39/0.37 & 0.7911 & 15.60/0.56 & 1.0515 & 17.09/0.43 \\
        \hline
        Mocycle-GAN + $\mathcal{L}_{PCP}$ & 0.6028 & 9.76/0.38 & 0.6539 & 16.07/0.57 & 0.9002 & 17.20/0.44 \\
        Mocycle-GAN + $\mathcal{L}_{EXS}$ & 0.6540 & 9.72/0.39 & 0.7834 & 16.58/0.57 & 0.9529 & 17.14/0.44 \\
        Mocycle-GAN + $\mathcal{L}_{INS}$ & 0.6238 & 9.67/0.37 & 0.6828 & 16.30/0.58 & 0.9632 & 17.21/0.44 \\
        \hline
        Recycle-GAN & 0.6306 & 9.41/0.37 & 0.5255 & 16.84/0.56 & 1.0609 & 16.34/0.43 \\
        \hline
        Recycle-GAN + $\mathcal{L}_{PCP}$ & 0.5387 & 9.61/0.39 & 0.4520 & 16.84/0.57 & 0.9134 & 17.01/0.43 \\
        Recycle-GAN + $\mathcal{L}_{EXS}$ & 0.6194 & 10.01/0.39 & 0.5259 & 16.88/0.58 & 1.0209 & 16.65/0.44 \\
        Recycle-GAN + $\mathcal{L}_{INS}$ & 0.5813 & 9.83/0.37 & 0.4610 & 16.98/0.58 & 0.9353 & 16.92/0.44 \\
        I2V-GAN w/o $\mathcal{L}_{PCP}$ & 0.6016 & 10.37/0.39 & 0.4987 & 16.24/0.59 & 1.0539 & 17.15/0.44 \\
        I2V-GAN w/o $\mathcal{L}_{EXS}$ & 0.5141 & 9.91/0.38 & 0.4688 & 16.65/0.58 & 0.9011 & 17.20/0.44 \\
        I2V-GAN w/o $\mathcal{L}_{INS}$ & 0.5311 & 10.28/0.39 & 0.4514 & 16.24/0.57 & 0.9996 & 17.19/0.44 \\
        \hline
    \end{tabular}
    \vspace{-1em}
\end{table*}

\subsection{Compared Approaches} 
We include the following state-of-the-arts and the most relevant unpaired translation methods for performance comparison: (1) CUT ~\cite{CUT} is an unpaired image translation method which utilizes contrastive learning. (2) PCSGAN ~\cite{PCSGAN} is an image-level I2V image translator base on GAN. (3) Cycle-GAN ~\cite{cycleGAN} pursuits an inverse translation at the image level to improve image translation performance. (4) Mocycle-GAN ~\cite{MocycleGAN} is a motion-guided Cycle-GAN for video translation that applies optical flow for motion estimation. (5) Recycle-GAN ~\cite{RecycleGAN} is the main baseline which leverages a recurrent temporal predictor to generate future frames and pursues a new cycle consistency across domains and time for unpaired video-to-video translation. (6) I2V-GAN is the proposed method in this paper. We compare the experimental results for each method and list the FID, PSNR and SSIM scores in Table ~\ref{tab:FID} and Table ~\ref{tab:PSNRSSIM}.

The sub-1 clip in IRVI has many drastic camera movements, which cause large semantic gaps between consecutive frames. While the camera is steady in other clips. In this situation, sub-1 is a hard case for video methods which consider temporal dependencies among frames. Image methods show advantages when frames are nearly individual, thus perform better than video ones. Our method performs better in most tasks, especially compared with video ones.

\begin{figure}[!t]
    \centering
    \includegraphics[height=6.5cm]{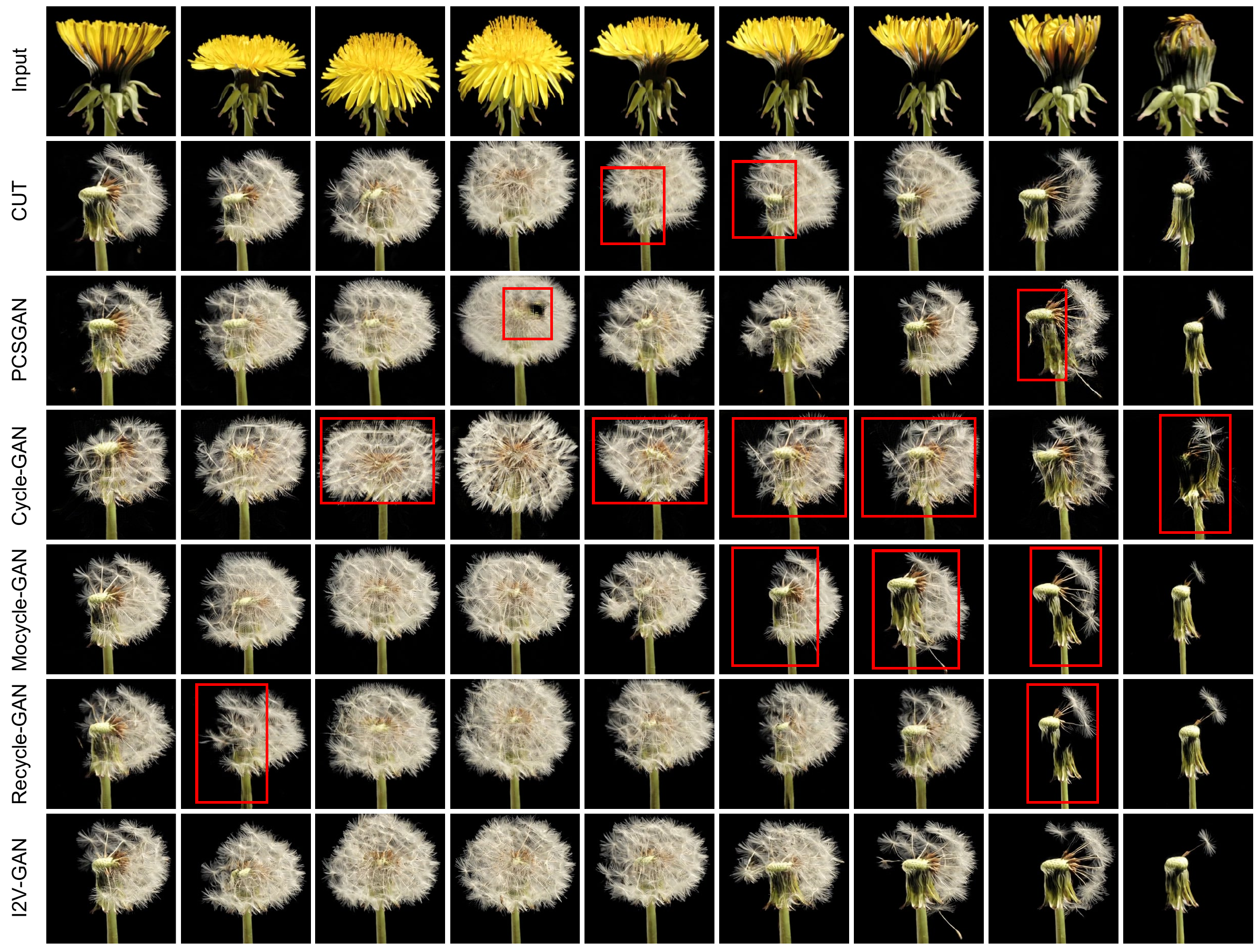}
    \vspace{-2em}
    \caption{Flower samples. Each row contains different life points of the flower, which generated by the methods noted on the left. Althougn some failures are not conspicuous at the image level, it is intolerable after combining them to a video since spatial-temporal consistency is not guaranteed.}
    \vspace{-2em}
    \label{fig:flowers_show}
\end{figure}

\subsection{Ablation Study and Further Investigation}
In this section, we further study the three proposed constraints: 1) perceptual cyclic loss, 2) external similarity loss, and 3) internal similarity loss. We add one of them to Mocycle-GAN and Recycle-GAN, then evaluate FID, PSNR and SSIM on IRVI traffic and monitoring all scene, as well as flower in the same setting. The ablation study implements by removing one of the three from I2V-GAN, which also evaluate FID, PSNR and SSIM. The experimental results in Table ~\ref{tab:ablation study} indicate our proposed constraints improve the performance.

\begin{table}[!ht]
    \caption{User study results.}
    \vspace{-1em}
    \label{tab:user study}
    \small
    \begin{tabular}{|m{25mm}<{\centering} | m{23mm}<{\centering} | m{23mm}<{\centering}|}
        \hline
        \textbf{Method} & \textbf{Realism} ${\uparrow}$ & \textbf{Fluency} ${\uparrow}$ \\
        \hline
        CUT & \textbf{7.74} / 10 & \textbf{5.21} / 10 \\
        PCSGAN & \textbf{5.79} / 10 & \textbf{6.94} / 10 \\
        Cycle-GAN & \textbf{4.57} / 10 & \textbf{4.41} / 10\\
        Mocycle-GAN & \textbf{4.14} / 10 & \textbf{7.23} / 10\\
        Recycle-GAN & \textbf{5.21} / 10 & \textbf{6.84} / 10\\
        I2V-GAN & \textbf{8.33} / 10 & \textbf{9.10} / 10\\
        \hline
    \end{tabular}
    \vspace{-2em}
\end{table}

\subsection{User study}
Since FID, PSNR and SSIM are not able to fully represent the real performance at the video level, we conduct an additional user study as shown in Table ~\ref{tab:user study}. We first select traffic translation result videos for each method in the same time period, as well as monitoring and flower. Then randomly arrange these videos without any information about which method generates them. After that, we ask 20 professional researchers to judge their realism and fluency scores from 1 to 10. The higher score indicates the better performance.

\begin{figure}[!t]
    \centering
    \includegraphics[height=5.6cm]{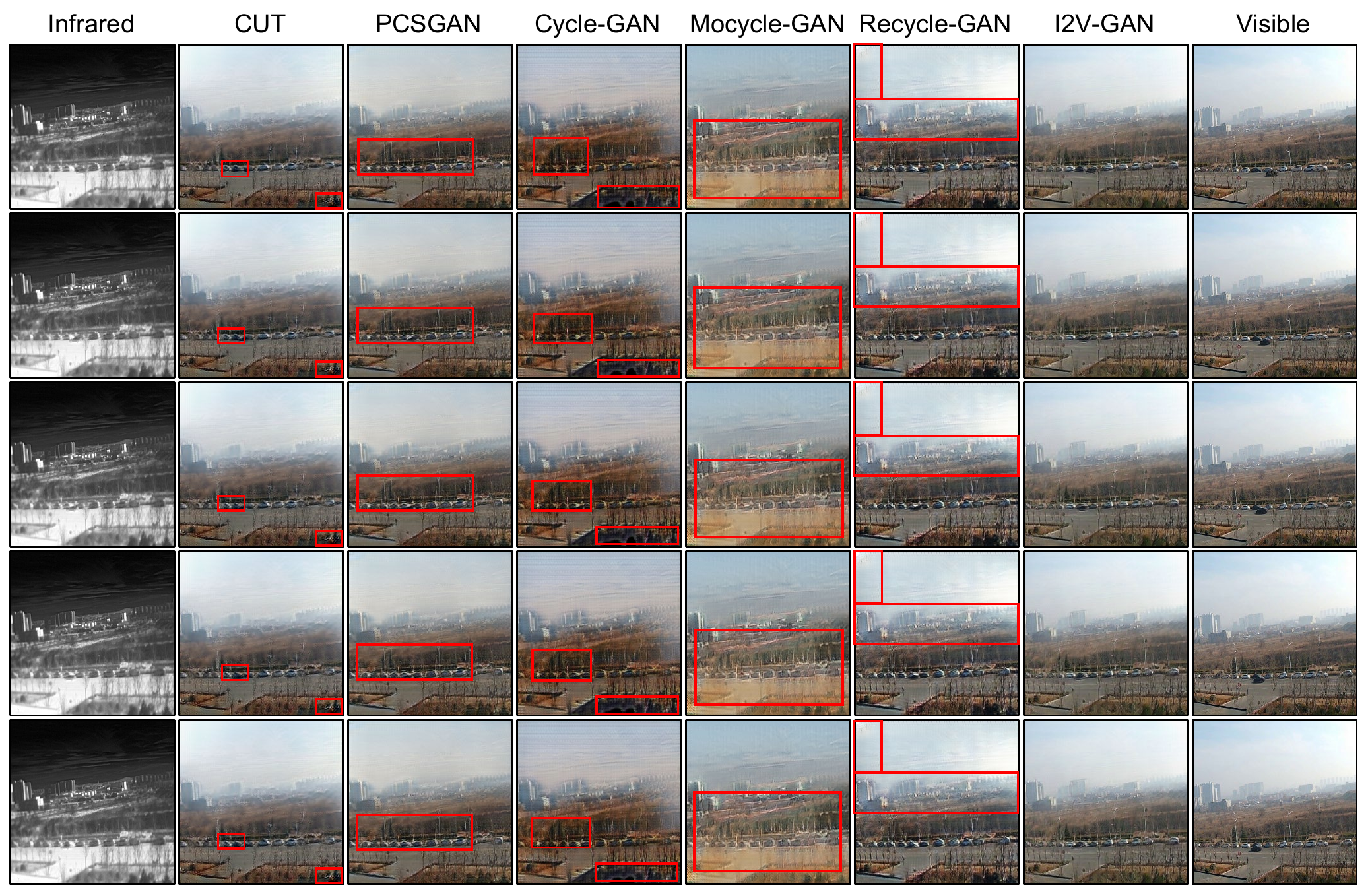}
    \caption{I2V monitoring samples, 5 consecutive frames from the top to bottom. The results of I2V-GAN are more realistic and fluent than the other compared methods.}
    \label{fig:frame}
\end{figure}


\section{Conclusion}
In this paper, we propose a novel infrared-to-visible video translation network named I2V-GAN, which could also be applied to other video translation tasks. Compared
with existing state-of-the-arts image-to-image and video-to-video translation methods, our method simultaneously improved the details and fluency for video translation in the unpaired setting. Extensive experiments show the efficacy of our proposal. In particular, we have compared detail effects for each part of our improvements in the ablation study. Moreover, additional user study from different perspectives demonstrate that I2V-GAN is more effective and suitable for real application scenarios.


\section*{Acknowledgements}
This work was supported by the National Natural Science Foundation of China (61902028).


\clearpage
\balance
\bibliographystyle{ACM-Reference-Format}
\bibliography{reference}

\end{document}